\documentclass{article}
\usepackage{ijcai23}
\usepackage[symbol]{footmisc}

\usepackage{times}
\usepackage{soul}
\usepackage{url}
\usepackage[hidelinks]{hyperref}
\usepackage[utf8]{inputenc}
\usepackage[small]{caption}
\usepackage{graphicx}
\usepackage{amsmath}
\usepackage{amsthm}
\usepackage{booktabs}
\usepackage{algorithm}
\usepackage{algorithmic}
\usepackage[switch]{lineno}

\usepackage{makecell}
\usepackage{amsfonts}
\usepackage{multirow}
\usepackage{dsfont}
\newcommand\x{x}

\title{ViT-CX: Causal Explanation of Vision Transformers}

\author{
Weiyan Xie $^1$
\and
Xiao-Hui Li $^2$\and
Caleb Chen Cao $^{1}$\And
Nevin L. Zhang $^1$
\affiliations
$^1$ The Hong Kong University of Science and Technology, China\\
$^2$ Huawei Technologies Co., Ltd, China
\emails
\{wxieai, cao, lzhang\}@ust.hk, \{lixiaohui33\}@huawei.com
}

\begin{document}
	
\maketitle

\begin{abstract}
	Despite the popularity of Vision Transformers (ViTs) and eXplainable AI (XAI), only a few explanation methods have been designed specially for ViTs thus far. They mostly use attention weights of the $[CLS]$ token on patch embeddings and often produce unsatisfactory saliency maps. This paper proposes a novel method for explaining ViTs called  {\em ViT-CX}. It is based on patch embeddings, rather than attentions paid to them, and their causal impacts on the model output. Other characteristics of ViTs such as causal overdetermination are also considered in the design of ViT-CX. The empirical results show that ViT-CX produces more meaningful saliency maps and does a better job revealing all important evidence for the predictions than previous methods. The explanation generated by ViT-CX also shows significantly better faithfulness to the model.  The codes and appendix are available at \url{https://github.com/vaynexie/CausalX-ViT}.
\end{abstract}

\begin{figure*}[h!]
	\centering
	\begin{tabular}{cccccc|cccccc}
		{\scriptsize {\tt  goldfish}}  & {\scriptsize CGW1} & {\scriptsize CGW2} & {\scriptsize TAM}& {\scriptsize ScoreCAM} & {\scriptsize ViT-CX}	 &{\scriptsize {\tt  dogsled}} & {\scriptsize CGW1} & {\scriptsize CGW2} & {\scriptsize TAM}& {\scriptsize ScoreCAM} & {\scriptsize ViT-CX}	\\
		\includegraphics[height=0.9cm,width=0.9cm]{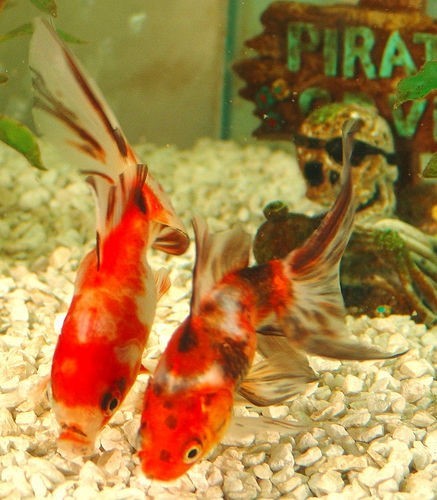} & 
		\includegraphics[height=0.9cm,width=0.9cm]{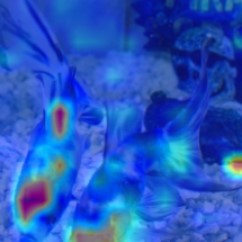} &
		\includegraphics[height=0.9cm,width=0.9cm]{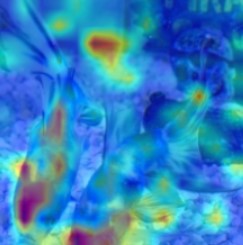} &
		\includegraphics[height=0.9cm,width=0.9cm]{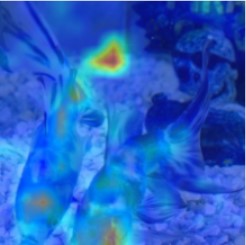} &
		\includegraphics[height=0.9cm,width=0.9cm]{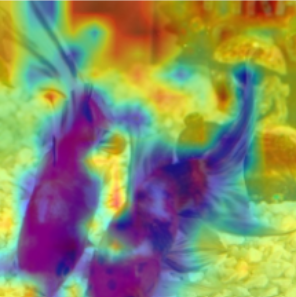} &
		\includegraphics[height=0.9cm,width=0.9cm]{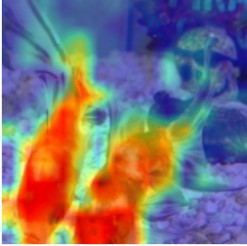}&
		\includegraphics[height=0.9cm,width=0.9cm]{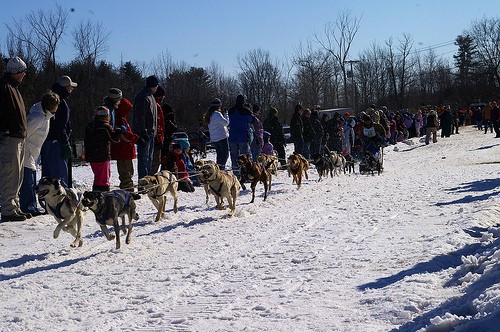} & 
		\includegraphics[height=0.9cm,width=0.9cm]{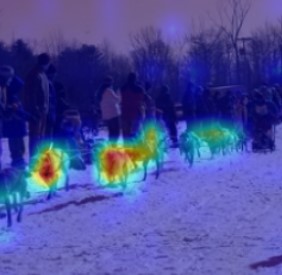} &
		\includegraphics[height=0.9cm,width=0.9cm]{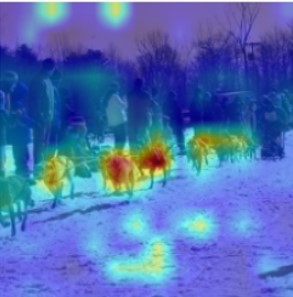} &
		\includegraphics[height=0.9cm,width=0.9cm]{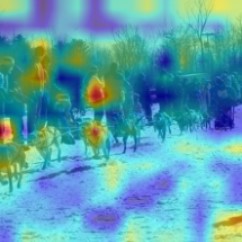} &
		\includegraphics[height=0.9cm,width=0.9cm]{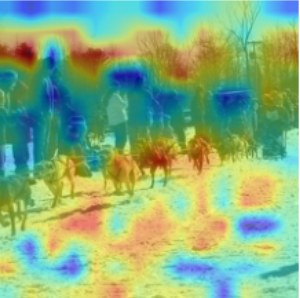} &
		\includegraphics[height=0.9cm,width=0.9cm]{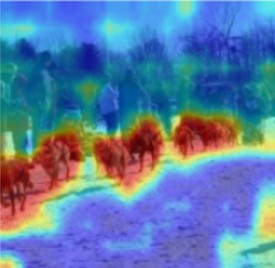} \\ \hline
		{\scriptsize Del$\downarrow$} &{\scriptsize 0.258} &{\scriptsize 0.355}&{\scriptsize 0.271}&{\scriptsize 0.532}&{\scriptsize \textbf{0.202}}&{\scriptsize Del$\downarrow$}&{\scriptsize 0.097}&{\scriptsize 0.147}&{\scriptsize 0.498}&{\scriptsize 0.698}&{\scriptsize \textbf{0.078}}\\
		{\scriptsize Ins$\uparrow$} &{\scriptsize 0.829} &{\scriptsize 0.833}&{\scriptsize 0.866}&{\scriptsize 0.553}&{\scriptsize \textbf{0.879}}&{\scriptsize Ins $\uparrow$}&{\scriptsize 0.827}&{\scriptsize 0.820}&{\scriptsize 0.692}&{\scriptsize 0.421}&{\scriptsize \textbf{0.884}}\\ \hline
		{\scriptsize {\tt  vine snake}} & {\scriptsize CGW1} & {\scriptsize CGW2} & {\scriptsize TAM}& {\scriptsize ScoreCAM} & {\scriptsize ViT-CX}	 &{\scriptsize {\tt  head cabbage}} & {\scriptsize CGW1} & {\scriptsize CGW2} & {\scriptsize TAM}& {\scriptsize ScoreCAM} & {\scriptsize ViT-CX}	\\
		
		\includegraphics[height=0.9cm,width=0.9cm]{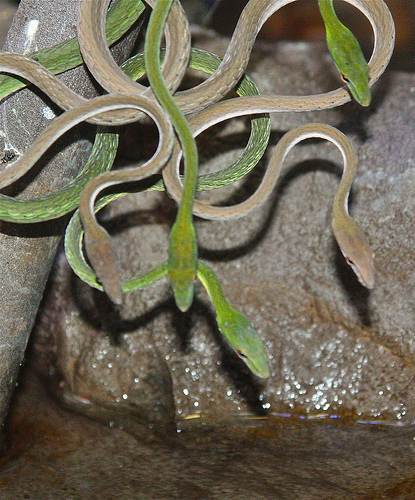}& 
		\includegraphics[height=0.9cm,width=0.9cm]{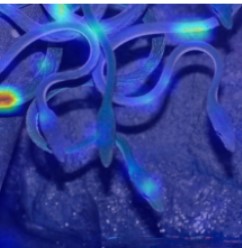} &
		\includegraphics[height=0.9cm,width=0.9cm]{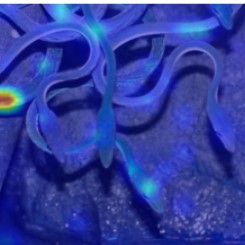} &
		\includegraphics[height=0.9cm,width=0.9cm]{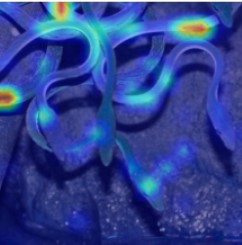}&
		\includegraphics[height=0.9cm,width=0.9cm]{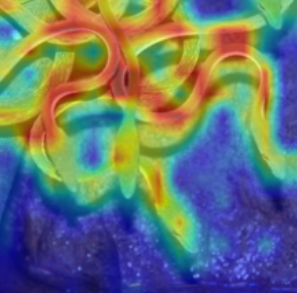}&
		\includegraphics[height=0.9cm,width=0.9cm]{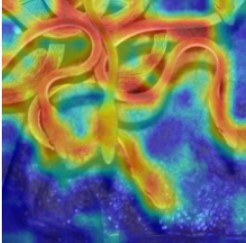}&
		\includegraphics[height=0.9cm,width=0.9cm]{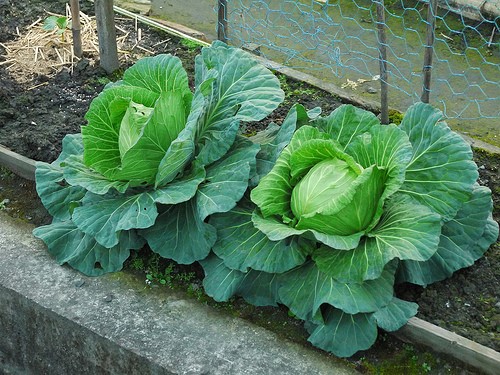}& 
		\includegraphics[height=0.9cm,width=0.9cm]{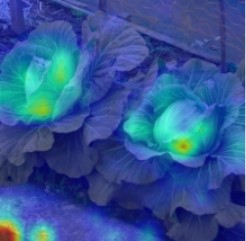} &
		\includegraphics[height=0.9cm,width=0.9cm]{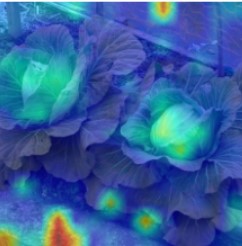} &
		\includegraphics[height=0.9cm,width=0.9cm]{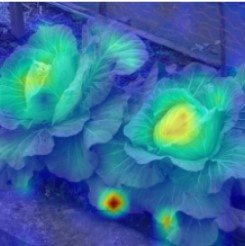}&
		\includegraphics[height=0.9cm,width=0.9cm]{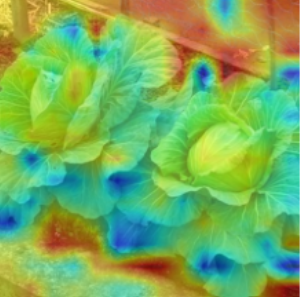}&
		\includegraphics[height=0.9cm,width=0.9cm]{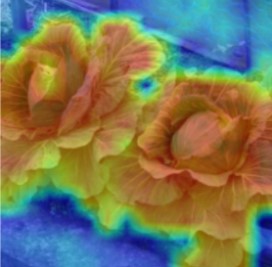} \\ \hline
		{\scriptsize Del$\downarrow$} &{\scriptsize 0.164} &{\scriptsize 0.122}&{\scriptsize 0.114}&{\scriptsize 0.108}& {\scriptsize \textbf{0.106} }&{\scriptsize Del$\downarrow$}&{\scriptsize 0.506}&{\scriptsize 0.598}&{\scriptsize 0.373}&{\scriptsize 0.634}&{\scriptsize  \textbf{0.351}}\\
		{\scriptsize Ins$\uparrow$} &{\scriptsize 0.410} &{\scriptsize 0.544}&{\scriptsize 0.337}&{\scriptsize 0.598}&{\scriptsize \textbf{0.603}}&{\scriptsize Ins$\uparrow$}&{\scriptsize 0.798}&{\scriptsize 0.780}&{\scriptsize 0.801}&{\scriptsize 0.535}&{\scriptsize \textbf{0.848}}\\ \hline
	\end{tabular}
	\caption{Explaining the predictions of ViT-B/16 on four images. The saliency maps by  ViT-CX are clearly more meaningful than those  by previous attention-based methods (CGW1, CGW2, TAM), highlighting all regions apparently important to predictions.  They are also more faithful to the model as measured by the deletion (Del) and insertion (Ins) AUC metrics. In contrast, the direct application of Score-CAM to ViTs can lead to nonsensical explanations (e.g., the {\tt goldfish} and {\tt head cabbage} example). }
	\label{score_cam_vit11}
\end{figure*}

\section{Introduction}

\begin{figure}[t]
	\centering
	\begin{tabular}{cc}
		
		\includegraphics[height=1cm,width=1cm]{goldfish2222/ILSVRC2012_val_00028713.JPEG}
		\begin{tabular}{ccccc}
			\includegraphics[height=0.9cm]{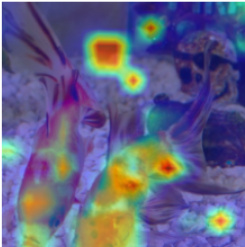}&
			\includegraphics[height=0.9cm]{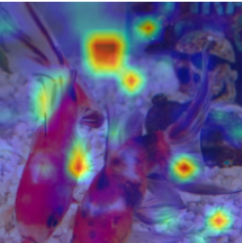}&
			\includegraphics[height=0.9cm]{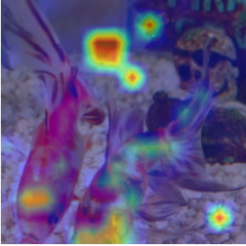}&
			\includegraphics[height=0.9cm]{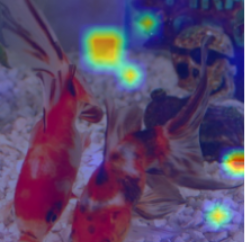}&
			\includegraphics[height=0.9cm]{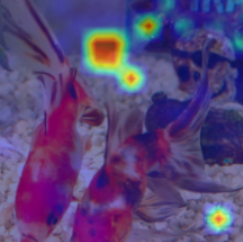}\\
			{\scriptsize (a.1)} & {\scriptsize (a.2)} &	{\scriptsize (a.3)} & {\scriptsize (a.4)} & {\scriptsize (a.5)}  \\
			\includegraphics[height=0.9cm]{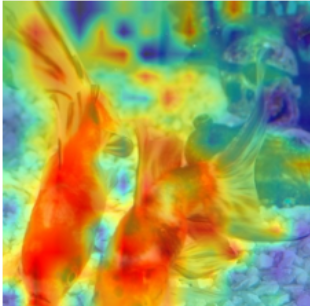}&
			\includegraphics[height=0.9cm]{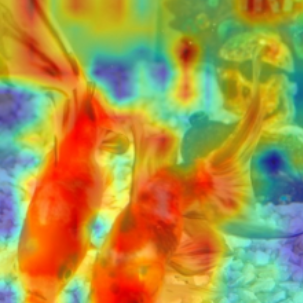}&
			\includegraphics[height=0.9cm]{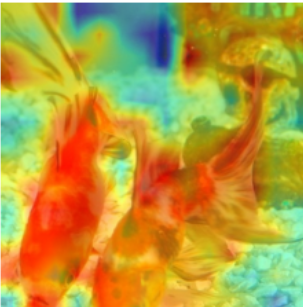} &
			\includegraphics[height=0.9cm]{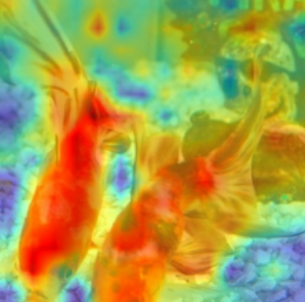} &
			\includegraphics[height=0.9cm]{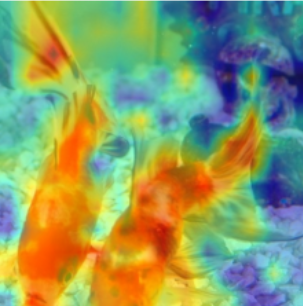}  \\
			{\scriptsize (b.1)} & {\scriptsize (b.2)} & {\scriptsize (b.3)} & {\scriptsize (b.4)} & 	{\scriptsize (b.5)} \\
			\includegraphics[height=0.9cm]{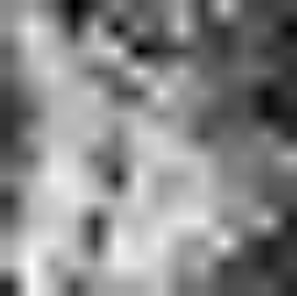}&
			\includegraphics[height=0.9cm]{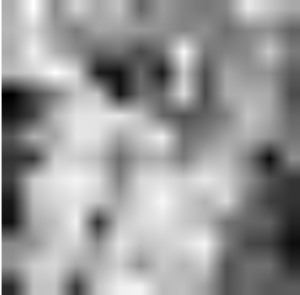}&
			\includegraphics[height=0.9cm]{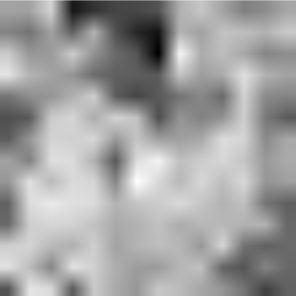} &
			\includegraphics[height=0.9cm]{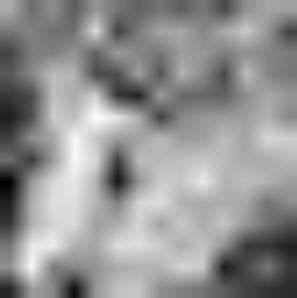} &
			\includegraphics[height=0.9cm]{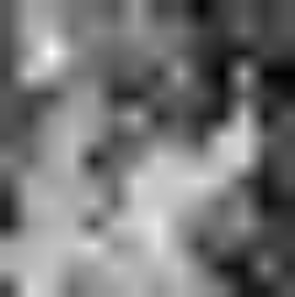}\\
			
			{\scriptsize (c.1)} & {\scriptsize (c.2)} & {\scriptsize (c.3)} & {\scriptsize (c.4)} &	{\scriptsize (c.5)}
		\end{tabular}
	\end{tabular}
	
	\caption{ViT feature maps (b.1 - b.5) are frontal slides of a 3D tensor made up of patch embedding vectors (as fibers). They are generally more
		meaningful than attention weight maps (a.1 - a.5), and they are used as ViT masks (c.1 - c.5) to generate explanations.
	}
	\label{vit1}
	\label{graph3}
\end{figure}

Vision Transformers (ViTs) are a new class of deep learning models that rival or even surpass the performance of convolutional neural networks (CNNs) on various vision tasks \cite{dosovitskiy2020image,carion2020end,liu2021swin}.  This paper is about explaining  the predictions by ViTs.  Several methods have been previously proposed for this task, namely CGW1 \cite{chefer2021transformer}, CGW2 \cite{chefer2021generic} and TAM \cite{yuan2021explaining}. Meanwhile, methods for explaining CNNs such as Grad-CAM \cite{selvaraju2017grad},  RISE \cite{petsiuk2018rise}, and Score-CAM \cite{wang2020score} can also be used to explain ViTs with minor adaptations. In this paper, we propose a novel method for explaining ViTs called {\em ViT-CausalX} or {\em ViT-CX} for short. Visual examples and experiment results show that ViT-CX  clearly outperforms previous baselines in terms of  faithfulness to model and interpretability to human users (Figure 1 and Table 1).

Previous ViT explanation methods are mainly based on attention weights of the class  token ($[CLS]$) on patch embeddings, or a combination of attention weights and class gradients. The use of attention weights for explaining NLP models has been extensively debated, and the general conclusion seems to point to the negative side \cite{jain2019attention,serrano2019attention,pruthi2020learning,bastings2020elephant}.   In ViTs, attention weights are concerned with the importance of patch embeddings to the $[CLS]$ token, but not the semantic contents of the embeddings themselves.  We conjecture that better explanations can be generated using the semantic contents of patch embeddings instead of the attentions paid to them.  ViT-CX is consequently developed.

ViT-CX is a specialized mask-based explanation method designed for ViT models. It generates masks by utilizing patch embeddings from a self-attention layer of a ViT model, which are arranged into a 3D tensor. The $(x,y)$-coordinates in the 3D tensor indicate the spatial information of the patch, while the $z$-coordinate represents its semantic content. By upsampling a frontal slice (with a fixed $z$) of the tensor to the input image size, a  {\em  ViT feature map} is produced. These feature maps (Figure \ref{vit1} (b.1 - b.5)) are more meaningful than the attention weight maps (Figure \ref{vit1} (a.1 - a.5)), and are used by ViT-CX to generate explanations. The method applies the feature maps as masks  (Figure \ref{vit1} (c.1 - c.5)) to the input image, calculates the causal impact score of the masks on the output, and combines the scores to generate saliency maps.

Other existing mask-based methods designed originally for CNNs include Occlusion \cite{zeiler2014visualizing}, RISE  \cite{petsiuk2018rise} and Score-CAM  \cite{wang2020score}.  Among them, Score-CAM, which uses CNN feature maps as masks, is ViT-CX's most similar counterpart for CNNs. However, there are three  technical issues that arise when adapting it directly to ViTs due to the characteristics of ViTs. We discuss these issues and include the solutions to them in ViT-CX.

The first issue is that applying a mask to an image might cause unintended artifacts when explaining ViTs. We consider this matter when calculating the causal impact scores of the masks. Second,  when using masks to make explanations, some pixels might be included in more masks than others, leading to {\em pixel coverage bias (PCB)}. PCB and its correction have been discussed in the context of random sampling \cite{petsiuk2018rise,sattarzadeh2021explaining}. However, this bias has not been widely addressed in mask-based methods for CNNs, including the Score-CAM. In this paper, we shows theoretically and empirically that PCB is a severe issue for ViTs due to the causal overdetermination. We also show that the PCB correction can significantly improve the quality of ViT explanations.

Third, mask-based methods usually require a large number of masks in order to generate high-quality explanations, which can lead to inefficient online performance, especially for ViTs which are usually heavier than CNNs.  To address this challenge, we empirically show that many ViT patch embeddings are similar, and propose clustering similar masks to significantly reduce the number of masks. We show that this strategy is effective for ViTs but does not work well for CNNs in general, as there is less variance among feature maps in ViTs than in CNNs. Note that one might suggest addressing the pixel coverage bias by using a large number of masks, which can make the explanation even more inefficient.   

In summary, we make the following contributions regarding ViT explanation in this paper:
\begin{itemize}
\item We propose to derive explanations for ViTs from the semantic contents of patch embeddings rather than attentions paid to them;
\item We develop a mask-based method for explaining ViTs that take into account the characteristics of ViTs, namely low variance of feature maps, strong shape recognition capability, and prevalence of causal overdetermination;
\item We empirically show that ViT-CX significantly outperforms previous baselines in terms both of the faithfulness to model and interpretability to human users.
\end{itemize}

\section{Related Work}

\subsection{Explanation Methods for Vision Transformers} 

The earliest methods for explaining ViT models are based on attention weights. All attention weights at an attention head can be reshaped and upsampled to the input size to form a saliency map. Rollout \cite{abnar2020quantifying} considers all heads from multiple layers and combines the corresponding attention maps to form one saliency map. Partial LRP \cite{voita2019analyzing} is similar to Rollout, except that it assigns different  weights to different heads, which are computed using Layer-wise Relevance Propagation (LRP) \cite{bach2015pixel}.  The saliency maps produced by Rollout and Partial LRP are not class-specific since the attention weights are class-agnostic. As such, those methods cannot be used to explain the reasons for particular output classes.

There are methods aiming to explain a particular output class.
CGW1 \cite{chefer2021transformer} is similar to Partial LRP, except that the gradients of the class score with respect to the heads are also considered, alongside LRP weights, when combining attention maps from different heads. In  CGW2 \cite{chefer2021generic}, the LRP weights are removed since they are found to be unnecessary. Transition Attention Map (TAM) \cite{yuan2021explaining} is similar to CGW2 except that simple gradients are replaced by integrated gradients \cite{sundararajan2017axiomatic}. Figure \ref{score_cam_vit11} shows several saliency maps produced by CGW1, CGW2 and TAM. They are clearly less satisfactory than those by ViT-CX. Moreover, attention weight-based methods are not applicable to ViTs \cite{liu2021swin,chu2021twins,zhang2022nested} that do not have a $[CLS]$ token, since  they utilize the attention maps between the  $[CLS]$ token and patch tokens. ViT-CX, on the other hand, relies on patch embeddings only and can be applied to a wider range of ViT variants.

\begin{figure*}[t]
	\centering
	\includegraphics[height=9.4cm]{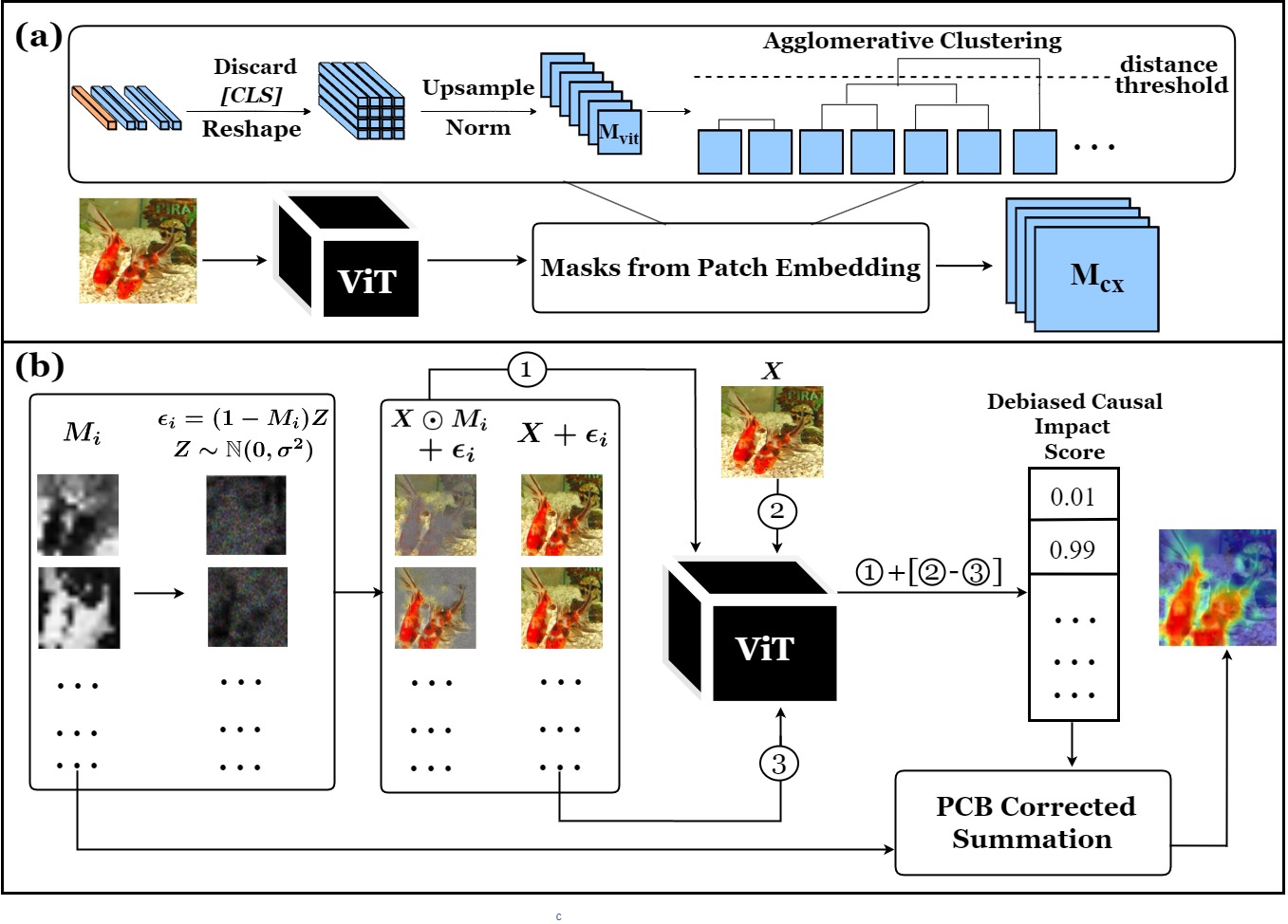}
	\caption{Overview of ViT-CX. (a) Mask Generation: A set of semantic masks is generated from the patch embeddings with agglomerative clustering. (b) Mask Aggregation: A saliency map is created by combining the masks using a debiased causal impact score based on the masked image with randomness to overcome artifacts. Pixel coverage frequencies are used in this step to correct pixel coverage bias.}
	\label{procedure1}
\end{figure*}

\subsection{Mask-based Explanation Methods} 

While there are only a few methods for explaining ViT models, a large number of methods have been proposed to explain CNN models. Mask-based explanation methods are one subclass. They generate an explanation based on a collection of masks $\mathbb{M}=\{M_1,\cdots,M_N\}$, where each mask $M_i$ is of the same size as the input image $X$, and its pixel values are between 0 and 1. A saliency map, as an explanation, is created by aggregating the masks weighted by {\em causal impacts}  of masks on the model output. Intuitively, we can think of a mask $M_i$ as a ``pixel team'', and the saliency value of a pixel is an aggregation of the causal impact scores of  ``teams" it is on. Occlusion \cite{zeiler2014visualizing}, RISE \cite{petsiuk2018rise} and Score-CAM  \cite{wang2020score} are typical mask-based explanation methods. 

 
 
In mask-based explanation methods, the causal impact of a mask  is usually measured by the class score ($f(\cdot)$) of the masked image $X \odot M_i$ on the target class $y$.  The saliency value of a pixel $x$ is determined by: 
\begin{equation}
	S(x) =  \sum_{i=1}^{N}f(y|X\odot M_i)M_i(x). \label{eq.score} 
\end{equation}
For visualization, the saliency values are normalized to interval $[0, 1]$ by $(S(x)- \min_x S(x))/(\max_x S(x)-\min_x S(x)$.

Score-CAM is a mask-based explanation method proposed to CNNs. It uses CNN feature maps as masks. One can apply it to ViTs by replacing CNN feature maps with ViT feature maps. However, this simple adaptation does not lead to quality explanations. ViT-CX improves it significantly by taking into account characteristics of ViTs.

ViT Shapley \cite{covert2022learning} is another mask-based explanation for ViTs proposed very recently. It trains a separate explainer (another ViT model) to estimate the Shapley Values. ViT Shapley has been evaluated only on small datasets (ImageNette and MURA) because the cost of training the explainer is high.
Its performance on large datasets such as ImageNet is difficult to assess. In addition, the explainer is a black-box and it introduces new opacity which might need further explanation.

\section{ViT-CX}

An overview of ViT-CX is shown in Figure \ref{procedure1}. ViT-CX follows the two-phase setting of mask-based explanation methods: mask generation followed by the mask aggregation.  In the first phase, a small set of semantic masks is generated from the patch embeddings in the target ViT model with a clustering algorithm applied to reduce the number and redundancy of masks (Section \ref{mask_generation.section}). In the second phase, we propose a \textbf{\em debiased causal impact score} to overcome the artifact bias (Section \ref{artifact.section}), and the final saliency map is obtained by \textbf{\em pixel coverage bias corrected summation}  ({{Section \ref{pcb.section}}}).

\subsection{Preliminaries}

In ViT models, an image $X \in \mathbb{R}^{H\times W\times C}$ is split into $N=HW/p^2$ patches,  with the $j$-th patch represented by a 2D vector $X_j\in \mathbb{R}^{(p\times p) \times C}$, where $H, W, C$ are the height, width, and the number of channels of the image, and $(p, p)$ is the spatial resolution of each patch. The patches are mapped to embeddings with $D$ dimensions via linear projection. The embeddings are fed into $L$ transformer blocks. Each block includes two modules: A {\em Multi-Head Self-Attention (MHSA) module}  and a {\em Multi-Layer Perceptron (MLP) module}. They yield new embeddings of the patches. We denote the patch embeddings at the output of transformer block $i$ as ${E}^{(i)} \in \mathbb{R}^{N_i\times D_i}$,  where $N_i$ is the number of patch tokens and $D_i$ is the feature dimension at that block. The $N_i$ and $D_i$ remain constant in vanilla ViT \cite{dosovitskiy2020image} as the computation proceeds from one block to another, and are gradually changed in the more recent ViTs \cite{wang2021pyramid,chu2021twins,liu2021swin} with  hierarchical structures.

\subsection{Mask Generation}	
\label{mask_generation.section}

\begin{figure}[t]
	\centering
	\includegraphics[height=2.8cm]{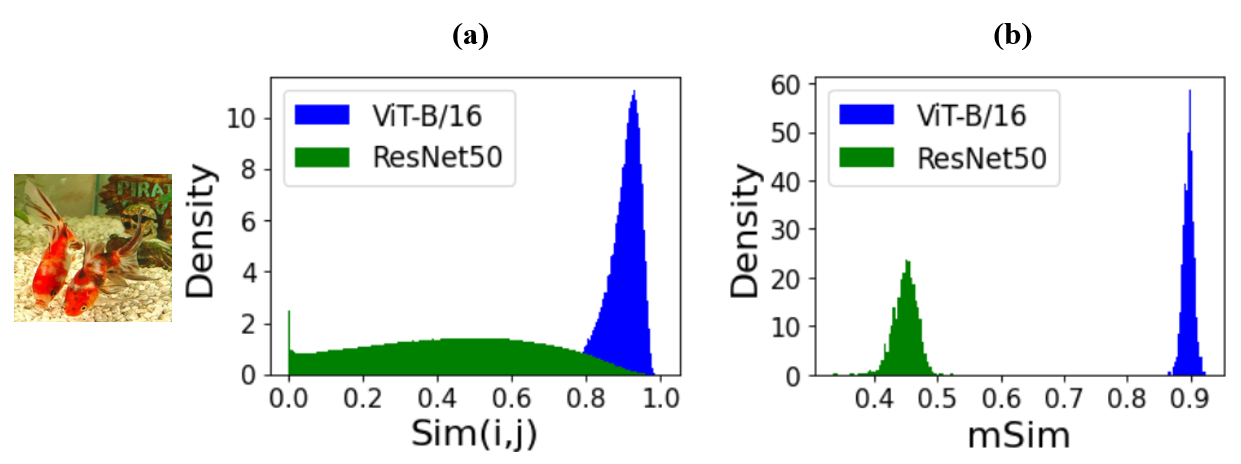}
\caption{ViT masks has lower variance then CNN masks. (a) Distribution of pairwise similarities ($Sim(i,j)$) between masks for the {\tt goldfish} image; (b) Distribution of mean pairwise mask similaritiy ($mSim$) for all images in the ImageNet validation set.}
	\label{overlap}
\end{figure}

\subsubsection{From ViT Feature Maps to ViT Masks}
We create the masks from the embeddings ${E}^{(i)}$ at the output of the attention module of a chosen transformer block $i$ (usually the last transformer block for the vanilla ViT and possibly other block(s) for other ViT models). The embeddings are first reshaped into a 3D tensor of size $\sqrt{N_i} \times \sqrt{N_i} \times D_i$. Each fiber in the tensor corresponds to the embedding of a patch, and 
the $(x, y)$-coordinates correspond to the spatial location of the patch in the input image. The frontal slices of the tensor are upsampled to the size of the input image, resulting in the {\em ViT feature maps}. The feature maps are subsequently normalized to the interval $[0, 1]$ to get {\em ViT masks}. A set of masks is built from those ViT masks, denoted as $\mathbb{M}_{vit} = \{M_1, \cdots, M_{D_i}\}$ where $M_j \in R^{H\times W}$ ($j=1,\cdots, D_i$). The number of masks in $\mathbb{M}_{vit}$ is $D_i$ ($D_i=768$ in ViT-B/16).  


\subsubsection{High Degree of  Redundancy  in ViT Masks} We observe a high degree of redundancy among the ViT masks. This is clear from Figure \ref{vit1} (c.1 - c.5). To quantify the level of redundancy among  ViT masks, we compute the pairwise cosine similarity between the mask $M_i$ and $M_j$:  $Sim(i,j)=\frac{M_{i}} {||M_{i}||}   \boldsymbol{\cdot}  \frac{M_{j}^{T}} {||M_{j}||} $.

The  mean pairwise cosine similarity $mSim$  is defined as: $mSim={\sum_{i=1}^{D_i}\sum_{j=i+1}^{D_i}{Sim}(i,j)}/[{D_i (D_i-1)/2}]$. There is a high average $mSim$  for the mask set  $\mathbb{M}_{vit}$ obtained from the last transformer block of ViT-B/16 - {0.92} \footnote{Average over the 50,000 images in ImageNet validation set.}, and the probability density distribution of the $mSim$ over images in ImageNet validation set is shown in Figure \ref{overlap} (b). Figure \ref{overlap} (a) shows the distribution of pairwise similarities $Sim(i,j)\ \ (j>i)$ for the {\tt goldfish} image, indicating most ViT masks of this image are close to each other. This points to the possibility of clustering similar ViT masks together to improve the explanation efficiency. 

In contrast, there is much more variance among CNN masks. This is clear from Figure \ref{overlap} where we also show the probability density distribution of masks $\mathbb{M}_{resnet}$ from a popular CNN, ResNet50 \cite{he2016deep}. Given that  most masks are not similar to each other in $\mathbb{M}_{resnet}$, applying the clustering on it can force the dissimilar masks to be grouped and cause a significant information loss. 


\subsubsection{Clustering the ViT Masks}
To reduce the redundancy in the mask set and improve the explanation efficiency, we merge the similar ViT masks. We use the agglomerative clustering algorithm \cite{mullner2013fastcluster,murtagh2014ward} which recursively merges data points with minimum pairwise distance. The pairwise distance of the mask $M_i$ and $M_j$  is measured by: 
\begin{eqnarray}
Distance(i,j)=1-\frac{M_{i}} {||M_{i}||}   \boldsymbol{\cdot} \frac{M_{j}^{T}} {||M_{j}||} \ .
	\nonumber	\label{cosine}
\end{eqnarray}
We stop the recursive merging based on a given distance threshold $\delta$ above which clusters will not be merged. Suppose the masks in $\mathbb{M}_{vit}$ is clustered into $K$ groups and each group is denoted as $\mathbb{M}_{vit}^{(k)}$ ($k=1, 2,\cdots, K$). Here $K$ is different for different images depending on the distribution of their ViT feature maps. We take the mean of the masks in each group to build the mask set $\mathbb{M}_{cx}$ used in ViT-CX:
\begin{eqnarray}
	&\mathbb{M}_{cx}=\{{M}_{1},{M}_{2},\cdots,{M}_{K}\},& \nonumber \\
	\text{where }{M}_k&= \frac{1}{|\mathbb{M}_{vit}^{(k)}|}\sum_{M\in \mathbb{M}_{vit}^{(k)}}M \ \text{, and } k=1,\cdots, K.& \nonumber
	\label{cosine1}
\end{eqnarray}
\noindent After the clustering, the number of masks decreases a lot ($K$ is {63} on average after clustering the previous  $\mathbb{M}_{vit}$ with $\delta=0.1$). The reduction in the number of masks reduces the number of ``pixel teams" that we need to investigate the causal impact for and thus can improve the online efficiency. At the same time, we show in Section 5 that reducing  the amount of  ``pixel teams" in this way has only a slight impact on the explanation quality.

\subsection{Artifacts and Debiased Causal Impact Score}

\begin{figure*}[h!]
	\centering
	\begin{tabular}{cc}
		\hspace{-0.2cm}
		\begin{tabular}{c}
			\ \\
			\includegraphics[width=1.5cm, height=1.5cm]{goldfish2222/ILSVRC2012_val_00028713.JPEG} \\
			{\scriptsize Input: \ P({\tt g})=0.998  }
		\end{tabular}
		&
		\begin{tabular}{c}
			\begin{tabular}{cc|cccc}
				&	{\scriptsize  ViT Feature Map}	 &	{\scriptsize  ViT Mask}	 &{\scriptsize  Masked Image} &{\scriptsize  Masked Image with Noise}&{\scriptsize  Image with Noise}	 \\
				&	 &	 {\scriptsize $M_i$}& {\scriptsize $X\odot M_i$} & {\scriptsize $X\odot M_i+\epsilon_i$} &{\scriptsize $X+\epsilon_i$}	 \\
					{{\bf Case (a)}}  &
				\includegraphics[height=1.02cm]{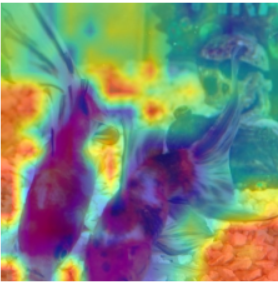} &
				\includegraphics[height=1.02cm]{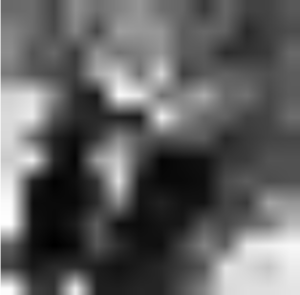} &
				\includegraphics[height=1.02cm]{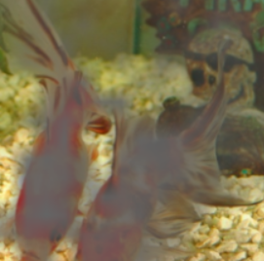} & 
				\includegraphics[height=1.02cm]{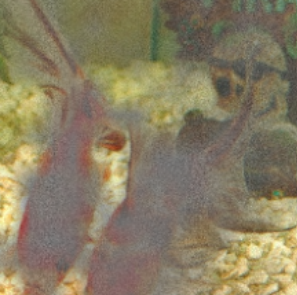}& 
				\includegraphics[height=1.02cm]{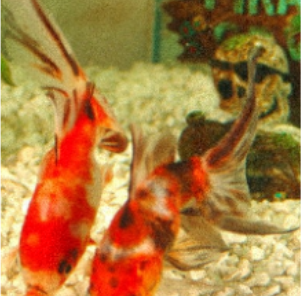}  \\  
				
			&		& &{\scriptsize P({\tt g})=0.973}  &{\scriptsize P({\tt g})=0.005} &{\scriptsize P({\tt g})=0.991}	\\ 
				&	&	& \multicolumn{3}{c}{{\scriptsize  Original CI = $\mathit{0.973}$, Debiased CI = $\mathbf{0.012}$}}	 \\ \hline  \multicolumn{2}{c|}{}& \multicolumn{4}{c}{}\\

				{{\bf Case (b)}} & 
				\includegraphics[height=1.02cm]{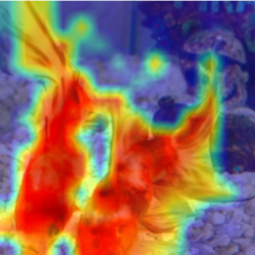} &
				\includegraphics[height=1.02cm]{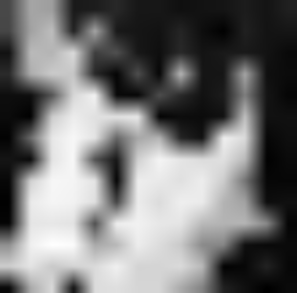} &
				\includegraphics[height=1.02cm]{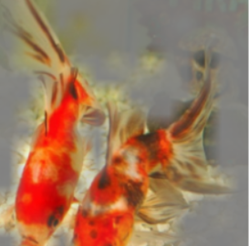} & 
				\includegraphics[height=1.02cm]{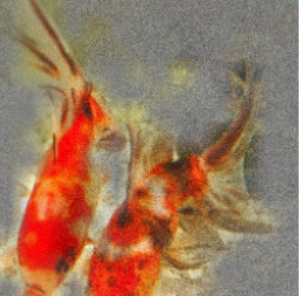} &
				\includegraphics[height=1.02cm]{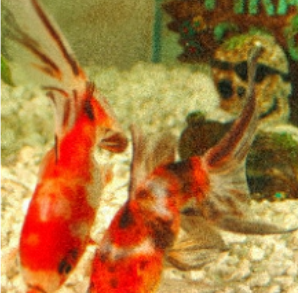}  \\ 
				
			&	  & &{\scriptsize P({\tt g})=0.994}  &{\scriptsize P({\tt g})=0.981} &{\scriptsize P({\tt g})=0.986}	\\ 
				  	& &	&	 \multicolumn{3}{c}{{\scriptsize  Original CI = $\mathit{0.994}$, Debiased CI = $\mathbf{0.993}$}}	 \\ 
		
			\end{tabular}
		\end{tabular}
	\end{tabular}
	\label{vit2}
\caption{Comparison of the original causal impact score (CI) and the proposed debiased causal impact score on the {\tt goldfish}  ({\tt g}) example. The original impact score is the prediction score of target class $y$ on the masked image - $f(y|X\odot M_i)$. The proposed debiased impact score based on masked images with randomness aims to address the artifact case (case (a)) where the masked pixels lead to misleading $f(y|X\odot M_i)$. Meanwhile the causal impact score in the normal case (case (b)) is not affected.}
	\label{vit2222}
\end{figure*}
\subsubsection{Artifactual Effects of  Masked Pixels} 

Considering the mask $M_i$ as a ``pixel team'', an implicit assumption in mask-based explanations is that only members of the ``pixel team" contribute to the causal impact score, while non-member pixels do not. 

In previous mask-based explanation methods, the causal impact score of a mask is usually measured by the prediction score of target class $y$ on the masked image -  $f(y|X\odot M_i)$, which can be problematic for ViTs. This problem arises from the violation of the implicit assumption. The masked pixels, as non-members of the ``pixel team", are at the same zero pixel value. They can co-create artifact that provides inferable information to the model and contributes strongly to the causal impact score, leading to the biased impact score.

Figure \ref{vit2222} case (a) shows an example of the artifact. The ViT feature map focuses on the background rather than the foreground, where the foreground pixels ({goldfish}'s pixels) share the least salient values. When the ViT mask is combined with the image, the foreground pixels are masked out. This erases the detailed features of the goldfish, such as texture and color. However, the masked pixels together create the shape of the goldfish in the masked image, resulting in an unreasonably high prediction score of {\tt goldfish} - 0.973. This phenomenon might relate to the stronger shape recognition ability of ViTs when making the inference, as pointed out by \cite{naseer2021intriguing,tuli2021convolutional}.

\subsubsection{Noise Addition to Correct Artifact Bias} 

To correct the artifact bias, we need to corrupt the information from the masked pixels with the same zero pixel values. This corruption can be achieved by adding random noise to these pixels. Therefore we propose to add the random noise in a soft way to the masked image:   $X\odot M_i+(1-M_i)Z$ , where $Z \in \mathbb{R}^{H\times W\times C}$ follows a Gaussian distribution  $\mathbb{N}(0, \sigma^2)$  with a {\bf small} standard deviation $\sigma$. Adding the noise based on the complement of mask values, i.e., $(1-M_i)$, allows only the distribution of masked pixels (non-member pixels) to be mainly affected while the distribution of pixels with the highest mask values (pixels in the ``team'') is minimally affected. In Figure  \ref{vit2222} (a), after the noise is added, the prediction score of {\tt goldfish} drops to 0.005. In case (b) where the goldfish body is preserved perfectly after masking, adding noises only makes the prediction score drop a tiny bit (0.994 $\rightarrow$ 0.981). 

\subsubsection{Debiased Causal Impact Score} Based on random noise addition, to reduce the effect of artifacts, we replace the term $f(y|X\odot M_i)$ in  Equation (\ref{eq.score}) with a debiased version of {\em causal impact score}:
\begin{eqnarray}
	&s(X, y, M_i)=&  \nonumber \\  &f(y|X\odot  M_i+\epsilon_i)+ [f(y|X)-f(y|X+\epsilon_i)], \nonumber &   \\   \\ 
	&\text{where }  \epsilon_i= (1-M_i)Z \in \mathbb{R}^{H\times W\times C}, Z\sim\mathbb{N}(0, \sigma^2).  \nonumber
	\label{eq.score.artifact11}
\end{eqnarray}

\noindent In Equation (2), the term $[f(y|X)-f(y|X+\epsilon_i)]$ is the drop on the prediction score of target class $y$ when the random noise is added to the unmasked image. We add this term to cancel the effect of the random noise addition and ensure the resulting scores purely reflect the effect of the masks $M_i$ on the image. The use of Equation (2) as the causal impact score when explaining ViTs is a good solution to the case of artifacts caused by the masked pixels, as the examples shown in Figure \ref{vit2222}.  The ablation study in Section 5 shows the effectiveness of this debiased score in more general cases.

\label{artifact.section}

\subsection{Pixel Coverage Bias and Its Correction}
\label{pcb.section}

Given the set of masks $\mathbb{M}_{cx}=\{M_1, M_2, \cdots, M_K\}$, the {\em coverage frequency} of a pixel $x$ is defined as: $$	\rho(x) = \frac{1}{K} \sum_{i=1}^K M_i(x).$$
\noindent  
Pixel coverage bias (PCB) refers to the phenomenon that different pixels might have different coverage frequencies. According to  Equation (\ref{eq.score}),
the saliency value of a pixel, before being normalized to $[0, 1]$, is the summation of the causal impact scores of the ``pixel teams'' (masks) of which it is a member.  Consequently, the more ``teams'' a pixel on, the higher its saliency value. This is clearly not justified.

\subsubsection{Adverse Effects of PCB} 
\begin{figure}[t]
	\centering
	\begin{tabular}{cccc}
		\includegraphics[height=1.1cm,width=1.1cm]{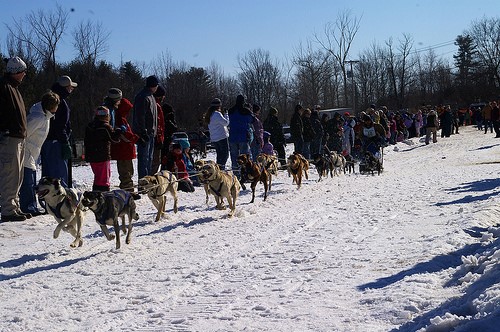} &
		\includegraphics[height=1.1cm]{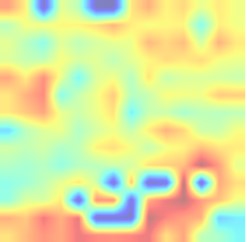} &
		\includegraphics[height=1.1cm]{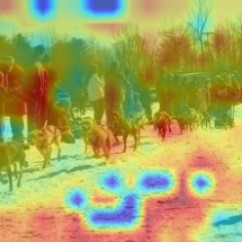}&
		\includegraphics[height=1.1cm]{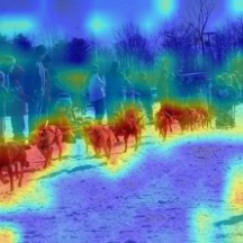} \\
		{\scriptsize (a) } &{\scriptsize (a.1)}&{\scriptsize (a.2)}&{\scriptsize (a.3)} 
	\end{tabular}
	\caption{Impact of PCB: The nonsensical saliency map (a.2) closely resembles the coverage frequency  map (a.1). After PCB correction, the map (a.3) becomes meaningful (target class - {\tt dogsled}). }
	\label{overdetermined}
	\label{fig.pcb11}
\end{figure}

Although pixel coverage bias is a common issue in mask-based explanation methods, it does not necessarily cause severe degradation in explanation quality.  However, it can severely degrade explanation quality in the cases of causal overdetermination. In those cases, the correct prediction can be made by various small patches of the input image \cite{white2021contrastive}, resulting that the causal impact score $s(X, y, {M}_i)$ of most masks is close to 1. To understand why PCB can cause undesirable explanation results in such cases, we let
$\mu=\frac1K \sum_{i=1}^K s(X,y,{M}_i)$  and $\beta_i=s(X,y,{M}_i)-\mu$, and divide the saliency score
$S(x)$ of a pixel into two parts:
\begin{eqnarray}
	S(x) & =&  \sum_{i=1}^{K} \beta_i {M}_i(x)+ \sum_{i=1}^{K} \mu {M}_i(x) \label{eq.score.rewrite0} \\ 
	&=&  \sum_{i=1}^{K} \beta_i {M}_i(x)+ \mu K \rho(x).
	\label{eq.score.rewrite}
\end{eqnarray}
In the  overdetermined cases, the $\mu$ (mean of impact scores) is close to 1, but $\beta_i$ is small for most $i$'s. Thus the second term, which is essentially the pixel coverage frequency, is much larger than the first term.  When normalized to the interval $[0, 1]$, the first term basically vanishes.  That leads to the saliency maps closely resemble the coverage frequency maps and fail to highlight areas important to the ViT prediction. One example of such case is given in Figure \ref{overdetermined}.

\subsubsection{Causal Overdetermination in ViTs} Causal overdetermination is common with ViTs. We find 20.95\% of the images in ImageNet validation set have a mean causal impact score $\mu$ greater than 0.9 when applying the mask $\mathbb{M}_{cx}$ to explain ViT/B-16. Among those images, the average variance of the score  is 0.036, which is small. They fit the features of high $\mu$ and low $\beta_i$ mentioned above. This finding is also consistent with \cite{naseer2021intriguing}'s observation that the class scores in ViTs are more robust to the removal of small patches in the input image than many popular CNNs.

\begin{table*}[t]
	\centering
	
	\begin{tabular}{c|ccc|ccc|ccc}
		\Xhline{1.2pt}	
		\multicolumn{1}{c}{}	& \multicolumn{3}{c|}{ViT-B} & \multicolumn{3}{c|}{DeiT-B}   & \multicolumn{3}{c}{Swin-B}                \\ \hline
		\multicolumn{1}{c|}{}	& Del $\downarrow$  &    Ins $\uparrow$ &PG Acc $\uparrow$&	  Del $\downarrow$  &    Ins $\uparrow$ &PG Acc $\uparrow$ &	  Del $\downarrow$ &    Ins $\uparrow$ &PG Acc $\uparrow$ \\ \hline
		{\bf \em ViT-CX} &\textbf{0.161} &\textbf{0.620}&\textbf{86.42\%}&\textbf{0.211}&\textbf{ 0.802} &\textbf{86.93\%} &\textbf{0.271} &\textbf{0.761} & \textbf{92.31\%}\\ 
		
		Number of Masks & \multicolumn{3}{c|}{Average: 63, Std: 11}& \multicolumn{3}{c|}{Average: 70, Std:12}& \multicolumn{3}{c}{Average: 95, Std:12} \\\Xhline{1.2pt}
		Rollout     &0.251  &0.517  & 60.91\%	 &0.406	 & 0.642& 35.70\% &--- &--- &--- \\
		Partial LRP   &0.239&0.499&66.52\% &0.349& 0.655 &61.25\%&--- &--- &--- \\
		CGW1   &0.201 & 0.542    &77.14\%   &0.286 &0.717& 70.54\%  &--- &--- &---    \\
		CGW2  & 0.209 & 0.549  & 70.94\% &0.271 &0.736 &70.54\% &--- &--- &---\\
		TAM &0.180&0.556&  \underline{77.87\%} &0.240& 0.747& 75.47\%&--- &--- &--- \\\hline 	
		{Occlusion}       & 0.291  & 0.571 & 64.75\% &	0.380 & \underline{ 0.801} & 59.51\% &0.448 & \underline{ 0.752} & 69.65\%\\ 
		{RISE}  &0.234&\underline{0.581}&73.30\%&0.366 & 0.759& 71.84\%  &0.416 &0.727 &75.07\%\\ 
		{Score-CAM} & 0.291& 0.471 & 48.89\% & 0.439 &0.576& 50.12\% &0.424  &0.641 & 69.65\% \\ \hline
		Grad-CAM  &0.212&0.456&50.45\%  &0.250 &0.743 & \underline{79.24 \%} & \underline{0.356} &0.693 & \underline{ 88.46\%} \\
		{Integrated-Grad}       &  0.184  & 0.263  &10.61\%  & 0.259 & 0.362 & 10.74\% &0.420 &0.483 &7.69\%\\
		{Smooth-Grad}       &  \underline{0.174}  &0.438 & 16.96\%   & \underline{0.231} &0.528 &31.05\%   &0.369 &0.505&14.52\%\\

		\Xhline{1.2pt}
	
	\end{tabular}
	\caption{Main experiment results: \textbf{Boldface} and \underline{underline} indicate best and second best performance, and `---' means not applicable.  ViT-CX significant outperforms all baselines in terms of the faithfulness metrics deletion (Del) and insertion AUC (Ins), and in terms of the interpretability metric Pointing Game Accuracy (PG Acc). The average number of masks ViT-CX used to explain an image is also provided.} 
	\label{robust_result1}
\end{table*}

\subsubsection{Correction for PCB} A simple way to correct for PCB is to divide the saliency value $S(x)$ by the coverage frequency $\rho(x)$. This 
results in the {\em corrected saliency value}:
\begin{eqnarray}
	S^c(\x) = \frac{S(x)}{\rho(x)}=  \sum_{i=1}^{K} s(X,y,{M}_i)\frac{{M}_i(\x)}{\rho(x)}\ , \label{eq.sc.1}
\end{eqnarray}
where $S^c(x)=0$ by definition when $\rho(x)=0$. Intuitively, 
the corrected saliency value of a pixel is the sum of the causal impact scores of the ``teams'' of which it is a member, divided by the number of ``teams'' it participates in. Similar to Equation (\ref{eq.score.rewrite}), we decompose 	$S^c(\x)$ into two parts:
\begin{eqnarray}
	S^c(x) = \sum_{i=1}^{K} \beta_i \frac{{M}_i(\x)}{\rho(x)}+ \mu K.
	\label{eq.score.corrected}
\end{eqnarray}
\noindent The second term is still much larger than the first term. However, it is a constant and does not depend on the pixel. When the saliency values $S^c(x)$ are normalized to the interval $[0, 1]$ for visualization, the influence of the second term is wholly eliminated. This is why meaningful saliency maps emerge in Figure \ref{overdetermined} after correcting for PCB. In Section 5.4, we show the correction improves the overall explanation quality greatly. 

This correction is also included in \cite{petsiuk2018rise} and \cite{sattarzadeh2021explaining}. Their motivation is to correct the bias caused by the finiteness of random sampling. We here provide another viewpoint on it and show its importance in causal overdetermination cases.

\section{Experiments}

\subsection{Evaluation Metrics} 
We evaluate ViT-CX following a protocol similar to how previous ViT explanation methods are evaluated, using the Deletion and Insertion AUC \cite{petsiuk2018rise}, Pointing Game \cite{zhang2018top} and  visual examples. This scheme is commonly used to evaluate explanation methods for CNNs.

\paragraph{Deletion and Insertion AUC:}  
The two metrics are about the {\em faithfulness} of an explanation (saliency map) to the target model, i.e., whether pixels with high saliency values are really important to the prediction \cite{petsiuk2018rise}.  Deletion AUC measures how fast the score of the target class drops as pixels are deleted from the image in descending order of the saliency values. Insertion AUC measures how fast the score increases when pixels are inserted into an empty canvas in that order. Smaller deletion AUC and larger insertion AUC indicate better faithfulness.

\paragraph{Pointing Game:}  
This metric is about the {\em interpretability} of an explanation, i.e., whether it provides qualitative understanding between input and output \cite{ribeiro2016should,doshi2017towards}. In Pointing Game, the saliency maps are compared with human-annotated bounding boxes. 
For each pair of saliency map and  bounding box, if the pixel with the highest saliency value falls inside the box, it is considered a hit. Otherwise it is considered a miss. The Pointing Game Accuracy is defined  as: $Acc={\#Hits}/({\#Hits+\#Misses})$. 


\begin{table*}[t]
	\centering
	
	\begin{tabular}{ccccccccc}
		\Xhline{1.2pt}

		\multicolumn{1}{c|}{} &{{\bf \em Masks}}	&  {{\bf \em Causal Impact}} &   {{{\bf \em PCB}}} & \multicolumn{1}{c|}{Number of Masks} 	& Del $\downarrow$  &    Ins $\uparrow$ &PG Acc $\uparrow$ & Average Time (s) \\ \hline
		\multicolumn{1}{c|}{{\bf \em ViT-CX}}
		&$\mathbb{M}_{cx}$  & $s(X, y, M_i)$  & $\surd$ & \multicolumn{1}{c|}{70 $ \pm $ 12}&\textbf{0.211} &0.802&\textbf{86.93\%}  & 1.15 $ \pm $ 0.15  \\ \Xhline{1.2pt}
		\multicolumn{1}{c|}{{Variant 1}} 	&$\mathbb{M}_{vit}$ & $s(X, y, M_i)$  & $\surd$ &\multicolumn{1}{c|}{768 $ \pm $ 0} & 0.232 & \textbf{0.810} &85.52\%  &8.23 $ \pm $ 0.03 \\
		\multicolumn{1}{c|}{Variant 2}	& $\mathbb{M}_{random}$  & $s(X, y, M_i)$  & $\surd$ &\multicolumn{1}{c|}{5000 $ \pm $ 0} &0.323 &0.734 & 75.12\% & 77.78 $ \pm $ 3.46  \\ \Xhline{1.3pt}

		\multicolumn{1}{c|}{{Variant 3}} &$\mathbb{M}_{cx}$   & $f(y|X\odot M_i)$  & $\surd$ &\multicolumn{1}{c|}{70 $ \pm $ 12}  &0.281 & 0.742 	 & 77.78\% 	 &0.98 $ \pm $ 0.12  \\	
	
		\multicolumn{1}{c|}{{Variant 4}} &$\mathbb{M}_{cx}$    & $s(X, y, M_i)$  & $\times$ & \multicolumn{1}{c|}{70 $ \pm $ 12}&0.303  & 0.727 	  &74.37\% 	  &1.12 $ \pm $ 0.14 \\
		\multicolumn{1}{c|}{{Variant 5}}  &$\mathbb{M}_{cx}$  	 & $f(y|X\odot M_i)$  & $\times$ &\multicolumn{1}{c|}{70 $ \pm $ 12}  & 0.339  &  0.686 	 & 67.56\%  &0.95 $ \pm $ 0.08\\\Xhline{1.3pt}
	\end{tabular}	
	\caption{Ablation study on ViT-CX (based on DeiT-B, GPU batch size=100). The analyzed components include: 1. {\bf\em Masks} - the mask set used in the explanation. $\mathbb{M}_{cx}$ is generated from ViT feature maps with reduced number by clustering and $\mathbb{M}_{vit}$ is the one without clustering. $\mathbb{M}_{random}$ is randomly generated;  2. {\bf\em Causal Impact} - the measure of casual impact of the mask. $s(X, y, M_i)$ is our proposed debiased causal impact score and  $f(y|X\odot M_i)$ is the prediction score of masked image; 3. {\bf\em PCB} - whether to apply the PCB correction.}
	\label{ablation.3}
	
\end{table*}

\subsection{Experiment Settings}

\noindent\textbf{Models and Dataset:} Three ViT variants are used in our experiments:  (1) ViT-B/16  \cite{dosovitskiy2020image}, the vanilla ViT;  (2) DeiT-B/16-Distill \cite{touvron2021training}, a improved version of the vanilla ViT with a distillation token; (3) Swin-B \cite{liu2021swin}, a hierarchical ViT. We use 5,000 images randomly selected from the ILSVRC2012 validation set \cite{deng2009imagenet}. {All experiments are run on  an Intel Xeon E5-2620 CPU and an NVIDIA 2080 Ti GPU.}

\paragraph{Hyper-parameters Setting:} To generate the masks $\mathbb{M}_{vit}$, we use feature maps from the last transformer block for ViT-B and DeiT-B, and choose those from the last block of the second to last stage for Swin-B; When clustering  on $\mathbb{M}_{vit}$ to generate the mask set $\mathbb{M}_{cx}$, the distance threshold $\delta$ is set to 0.1 for ViT-B and DeiT-B, and set to 0.05 for Swin-B; The standard deviation $\sigma$  of the Gaussian noise $\epsilon_i$ is set to 0.1.

\paragraph{Baselines:} In this section, we compare ViT-CX with three groups of baselines: (a) Five attention weights-based methods, namely Rollout \cite{abnar2020quantifying}, Partial LRP \cite{voita2019analyzing}, CGW1 \cite{chefer2021transformer}, CGW2 \cite{chefer2021generic}, and TAM \cite{yuan2021explaining};  (b) Three mask-based methods, namely Occlusion \cite{zeiler2014visualizing}, RISE \cite{petsiuk2018rise} and Score-CAM \cite{wang2020score}; (c) Three gradient-based methods, namely  Grad-CAM \cite{selvaraju2017grad}, Integrated-Grad \cite{sundararajan2017axiomatic} and Smooth-Grad \cite{smilkov2017smoothgrad}. In addition, Appendix D provides a comparison between ViT-CX and ViT Shapley, a mask-based explanation method for ViTs introduced in a recent study by \cite{covert2022learning}.

\subsection{Results}

The main results are in Table \ref{robust_result1}.

\paragraph{Faithfulness:} ViT-CX has the lowest deletion AUC values across the board,  being 10\% lower than the next best. ViT-CX also enjoys the highest insertion AUC values in all cases. Those indicate that ViT-CX is more faithful to the target models than all baselines.

\paragraph{Interpretability:} ViT-CX enjoys significantly higher Pointing Game accuracy than the baselines in all cases. This implies that the explanations of ViT-CX are more consistent with human-annotated bounding boxes. As a supplement to quantitative metrics, four visual examples have been shown in Figure 1 and more examples are given in Appendix A.

\paragraph{Computation Cost:} ViT-CX uses less than 100 masks on average  to explain an image. The computation cost is greatly reduced compared to previous mask-based methods.  

\paragraph{Sanity Check:}
As a causal method, ViT-CX is sensitive to the changes in model parameters and 
passes the sanity check \cite{adebayo2018sanity}. See {Appendix B} for details.

\paragraph{Localization:} As a training-free method, ViT-CX shows comparable localization performance to recently proposed ViT-based weakly supervised object localizers that require extra modules and training. The details are in {Appendix C}.


\subsection{Ablation Study}

The collection of masks used in the explanation, the causal impact score, and whether the PCB correction is applied are varied to study the effect of different components in ViT-CX. The results are in Table \ref{ablation.3}.

Comparison of ViT-CX with Variant 1 shows that  clustering on the mask set $\mathbb{M}_{vit}$ reduces the number of masks from 768 to 70 on average, and the mean time to explain an image is reduced to $\sim$1 second. Meanwhile the explanation quality is mainly remained unaffected. Additionally, ViT-CX outperforms explanations generated with random masks (Variant 2) in terms of explanation quality and efficiency. 


Comparisons with Variants 3-5 emphasize the significance of addressing pixel coverage bias and using a debiased causal impact score in ViT-CX to avoid artifact effects. Without these steps, the explanation quality is largely affected. These results suggest that the clustering of masks, the correction for PCB and the debiasing of causal impact scores are crucial components of ViT-CX. A more detailed sensitivity analysis of hyperparameters in ViT-CX  can be found in Appendix F.

\section{Conclusion}

Previous attention weights-based and mask-based explainers have not been able to consistently provide satisfactory explanations for ViTs. ViT-CX, a specially designed mask-based explainer for ViTs, addresses the issues of low explanation efficiency, misleading causal impact scores caused by artifacts, and pixel coverage bias of masks. Our solutions to these issues are demonstrated to lead to high-quality explanations for various ViT image classifiers. Future work could involve extending ViT-CX concepts to explain ViT models for other tasks like object detection and segmentation.


\section*{Acknowledgments}
We thank the deep learning computing framework MindSpore (\url{https://www.mindspore.cn}) and its team for the support on this work. Research on this paper was supported in part by Hong Kong Research Grants Council under grant 16204920. Weiyan Xie was supported in part by the Huawei PhD Fellowship Scheme. We thank Prof. Janet Hsiao, Yueyuan Zheng, Luyu Qiu, and Yi Yang for valuable discussions.

\section*{Contribution Statement}
Weiyan Xie and Xiao-Hui Li contributed equally to this work. This work is done when Caleb Chen Cao was in Huawei Research Hong Kong.

\bibliographystyle{named}
\bibliography{ref.bib}

\end{document}